\documentclass[10pt,twocolumn,letterpaper]{article}

\usepackage[pagenumbers]{cvpr}      
\usepackage{amsmath}
\usepackage{amssymb}
\usepackage{mathtools}
\usepackage{amsthm}
\usepackage{braket}
\usepackage{wrapfig}
\usepackage{multirow}

\usepackage[dvipsnames]{xcolor}

\definecolor{cvprblue}{rgb}{0.21,0.49,0.74}
\usepackage[pagebackref,breaklinks,colorlinks,citecolor=cvprblue]{hyperref}

\usepackage[customcolors]{hf-tikz}
\usepackage{amsmath}

\usepackage{amsmath,amsfonts,bm}

\def\eqref#1{equation~\ref{#1}}

\def\1{\bm{1}}

\def\rq{{\textnormal{q}}}

\def\ry{{\textnormal{y}}}

\DeclareMathAlphabet{\mathsfit}{\encodingdefault}{\sfdefault}{m}{sl}
\SetMathAlphabet{\mathsfit}{bold}{\encodingdefault}{\sfdefault}{bx}{n}

\def\gQ{{\mathcal{Q}}}

\def\gT{{\mathcal{T}}}

\def\gX{{\mathcal{X}}}
\def\gY{{\mathcal{Y}}}

\def\sQ{{\mathbb{Q}}}

\usepackage{mathtools}
\usepackage{multirow}
\usepackage{multicol}
\usepackage{booktabs}
\usepackage{subfigure}

\DeclareMathOperator*{\compose}{c}

\DeclareMathOperator*{\genprompt}{GoodQ}
\DeclareMathOperator*{\conditionprompt}{CondQ}

\def\paperID{12586} 
\def\confName{CVPR}
\def\confYear{2024}

\title{Good Questions Help Zero-Shot Image Reasoning}

\author{%
  Kaiwen Yang\textsuperscript{1} \quad Tao Shen\textsuperscript{2} \quad Xinmei Tian\textsuperscript{1} \quad  Xiubo Geng\textsuperscript{3} \quad  \\ Chongyang Tao\textsuperscript{3} \quad Dacheng Tao\textsuperscript{4} \quad Tianyi Zhou\textsuperscript{5}\\
University of Science and Technology of China\textsuperscript{1}; University of Technology Sydney\textsuperscript{2}\\Microsoft Corporation\textsuperscript{3}; The University of Sydney\textsuperscript{4}; University of Maryland, College Park\textsuperscript{5} \\ 
\href{https://github.com/kai-wen-yang/QVix}{\texttt{https://github.com/kai-wen-yang/QVix}}
}

\begin{document}
\maketitle
\begin{abstract}

Aligning the recent large language models (LLMs) with computer vision models leads to large vision-language models (LVLMs), which have paved the way for zero-shot image reasoning tasks. However, LVLMs are usually trained on short high-level captions only referring to sparse focus regions in images. Such a ``tunnel vision'' limits LVLMs to exploring other relevant contexts in complex scenes. 
To address this challenge, we introduce Question-Driven Visual Exploration (QVix), a novel prompting strategy that enhances the exploratory capabilities of LVLMs in zero-shot reasoning tasks. QVix leverages LLMs' strong language prior to generate input-exploratory questions with more details than the original query, guiding LVLMs to explore visual content more comprehensively and uncover subtle or peripheral details. QVix enables a wider exploration of visual scenes, improving the LVLMs' reasoning accuracy and depth in tasks such as visual question answering and visual entailment. Our evaluations on various challenging zero-shot vision-language benchmarks, including ScienceQA and fine-grained visual classification, demonstrate that QVix significantly outperforms existing methods, highlighting its effectiveness in bridging the gap between complex visual data and LVLMs' exploratory abilities. 
\end{abstract}    
  \vspace{-1.em}
\section{Introduction}
\label{sec:intro}






\begin{figure}[t]
  \centering
   \includegraphics[width=1\linewidth]{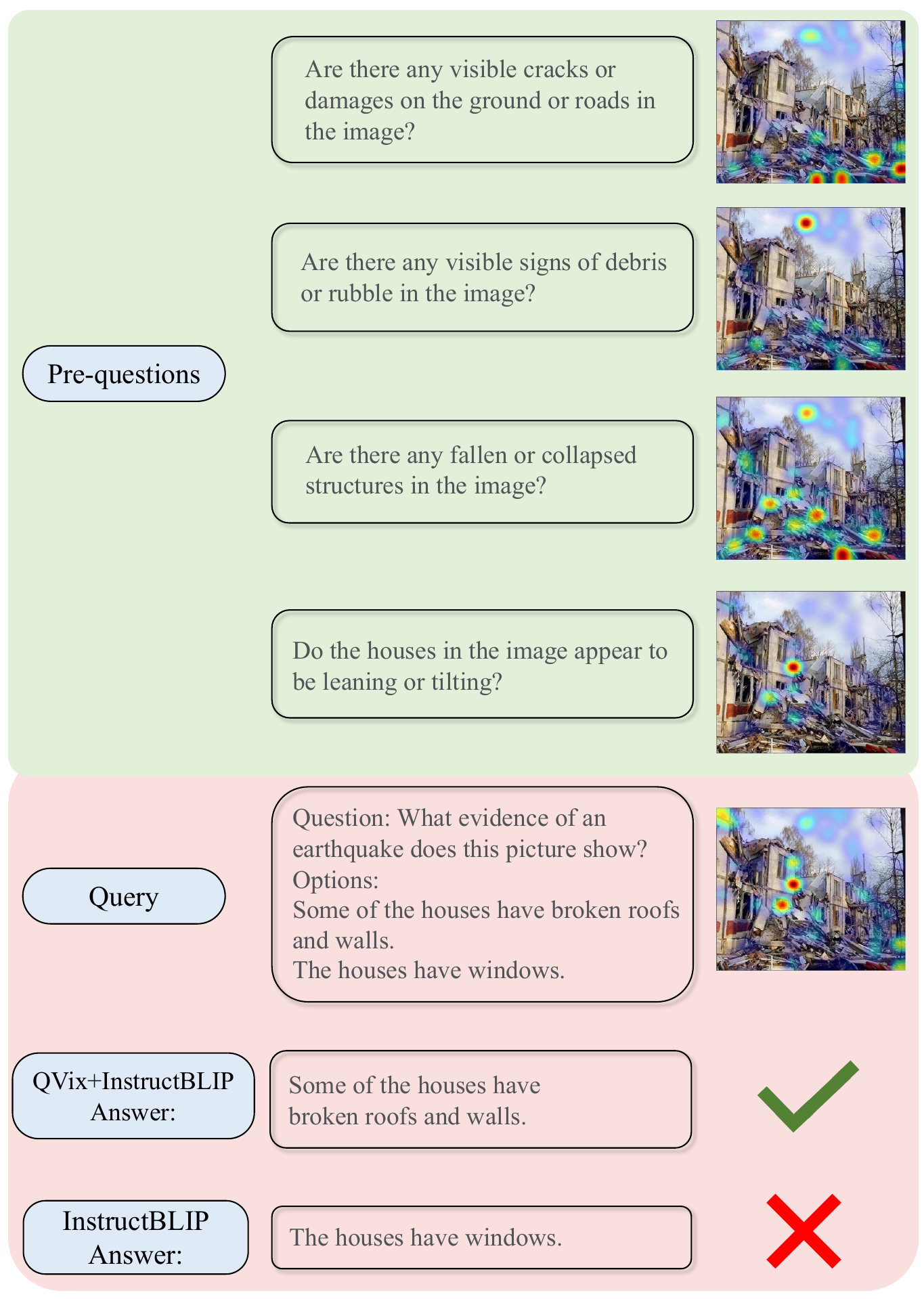}

   \caption{\footnotesize Cross-attention map between an image and QVix pre-questions vs. the original query. QVix pre-questions help LVLMs explore contextual regions related to the original query. The attention is generated by the Q-former in InstructBLIP~\cite{instructblip}.}
   \label{fig:instruct}
    \vspace{-1.2em}
\end{figure}

In the last two years, large language models (LLMs), e.g., GPT4~\cite{openai2023gpt4}, LLaMA~\cite{touvron2023llama}, have been showing extraordinary capability in natural language processing and understanding tasks, such as question answering~\cite{antol2015vqa}, machine reasoning~\cite{dai2019bridging}, math solving~\cite{zong2023solving}, code programming~\cite{xu2022systematic}, as well as open-ended response generation. Beyond natural language, it is more than exciting to adapt more modalities into large language models, upon an inspiration that an LLM pre-trained on trillions of language tokens serves as a reasoning brain while other modalities aligned to the LLM make the best of the reasoning ability. Among this, computer vision, e.g., the modality of images, is particularly intriguing due to its extensive range of applications and the richness of aligned vision-language (VL) data, leading to large vision language models (LVLMs) \cite{instructblip, liu2023improved, zhu2023minigpt, liu2023visual, li2023blip2}. 
Built upon a well-trained visual encoder and an LLM, they are trained on large-scale image-text dataset to align the output of a visual model to the input space of a Large Language Model (LLM), and lastly, fine-tuned on a relatively small-scale instruction-following image-text dataset to enable the model to understand and execute task-specific instructions.

The integration of computer vision with LLMs opens up new frontiers in zero-shot image reasoning tasks, e.g., Visual Question Answering (VQA) \cite{antol2015vqa} and Visual Entailment (VE) \cite{xie2019visual}, where the model is expected to understand and reason about images without explicit prior training on them. 
Intuitively, solving these tasks requires an LVLM to not only understand the content of images but also to effectively integrate this understanding with its natural language reasoning capabilities. 
However, except for close-source LVLMs like GPT-4V~\cite{yang2023dawn}, most open-source ones, e.g., InstructBLIP~\cite{instructblip} and MiniGPT-4~\cite{zhu2023minigpt}, adapt the visual encoder to the LLM usually by aligning an informative image with a sparse caption. Such an asymmetric alignment makes the resulting LVLM prone to focus only on the image regions that can be grounded into the associated natural language question, regardless of other relevant contexts. This behavior, which we named ``\emph{tunnel vision}'', could be beneficial in tasks where the image-text alignment is clear and direct. It, however, becomes a limitation in scenarios where the visual data contains nuances or information that could elicit rich prior knowledge from LLMs, such as ScienceQA~\cite{lu2022learn} and fine-grained image classification~\cite{parkhi2012cats, WahCUB_200_2011}.

For example, as shown in Fig.~\ref{fig:instruct}, given just the original query, the LVLM only focuses on limited areas in the image that are related to several words contained in the input query, such as trees, and fails to pay attention to other areas potentially relevant to the query.
Consequently, these LVLMs often overlook the deeper, context-driven insights that could be gleaned from a more holistic interpretation of the visual content, limiting their effectiveness in complex image reasoning tasks.

Previous reasoning methods focus on decomposing the original question either implicitly (e.g., chain-of-thought (CoT) \cite{wei2022chain, wang2022self}) or explicitly (e.g., least-to-most \cite{Zhou2023Least, wu2023visual}). 
Nonetheless, these methods still fall short in addressing the \emph{tunnel vision} issue, as they largely depend on the initial framing of the question, which often biases the exploration towards certain aspects of the image while neglecting other relevant contexts. Additionally, question decomposition approaches usually fail to adaptively explore the visual scene based on the evolving context, thereby limiting their ability to uncover nuanced or peripheral details that are not directly hinted at in the original question. This results in a constrained understanding of the visual content, impeding the LVLMs' ability to fully leverage their language reasoning capabilities in diverse and complex visual scenarios.


This challenge motivates us to propose a new prompting framework, while an open research question is how to make the best of input-related prior knowledge in parametric LLMs, which can guide LVLMs to perform a more nuanced and comprehensive analysis of the visual data. 
To this end, we present a framework, dubbed Question-Driven Visual Exploration (QVix), for the purpose of enlarging vision scope in LVLMs for zero-shot image reasoning.  
With an inspiration that question-asking is a powerful tool for widening attention, this approach relies on the hypothesis that well-crafted questions can stimulate the LVLM to delve beyond surface-level details, enriching more subtle or peripheral visual cues that might otherwise be neglected. QVix is to elicit prior knowledge from LLMs by allowing them to ask good input-exploratory questions that drive the LVLM to explore visual information more comprehensively when performing reasoning. As a result, such a comprehensive image exploration can help the model understand the image-associating input query more thoroughly and answer it more precisely. Note that ``good'' doesn't mean the asked questions are \emph{optimal} but just \emph{good} to provide extra prompts to guide the LVLM towards a more thorough examination of the visual data.

For instance, as shown in Fig.~\ref{fig:instruct}, the main question posed to the LVLM is ``\textit{What evidence of an earthquake does this picture show?
}''. By referring to the main question directly, the LVLM might focus solely on the most prominent objects in the main question, such as ``\textit{houses}'', neglecting key elements. In contrast, with QVix, the LLM generates additional, insightful pre-questions that prompt a more thorough exploration. Questions like ``\textit{Are there any fallen or collapsed structures in the image?
}'' or ``\textit{Are there any visible cracks or damages on the ground or roads in the image?
}'' guide the LVLM to consider parts of the image it might have otherwise ignored. As a result, the LVLM attends to ``\textit{broken roofs}'' and ``\textit{walls}'' of the house, which is the strongest evidence of an earthquake. 

Hence, by engaging the LVLM in a more exhaustive visual exploration, QVix is able to enhance its interpretative depth and accuracy in zero-shot image reasoning tasks. This brand-new prompting method effectively bridges the gap between the visual data's complexity and the LVLM's interpretative capabilities, ensuring more comprehensive and contextually aware image analysis.

Empirically, we evaluate our method on several challenging VL tasks in a zero-shot setting in both quantitative and qualitative ways, including ScienceQA~\cite{lu2022learn}, image classification, and Visual Entailment (SNLIVE)~\cite{xie2018visual}. In contrast to our competitors, our method can outperform the best on all the tasks. On the challenging ScienceQA, our method outperforms the baseline by $5.7\%$ average accuracy.

\section{Related Work}
\label{sec:formatting}

\paragraph{Large Visual-language Models (LVLMs).}
In order to leverage the rich prior knowledge and complex reasoning ability of the recent well-established LLMs~\cite{radford2018improving, touvron2023llama}, an increasing number of studies have been proposed to make visual modality available to pre-trained LLMs~\cite{zhu2023minigpt, liu2023improved, liu2023visual}. Typically, these LVLMs follow a two-stage training manner. In the first stage, these LVLMs introduce a projection layer \cite{liu2023visual} or a Q-former network \cite{zhu2023minigpt, liu2023improved} and train it on large-scale image-text pairs dataset to align the visual feature to the input space of LLM. In the second stage, in order to enable LVLMs to follow users' order to generalize to unseen tasks, some methods finetune LVLMs using extra instruction-following data. LLaVA~\cite{liu2023visual} constructs a vision-language conversational dataset called LLaVA-150k generated by GPT-4~\cite{openai2023gpt4} and finetune the projection layer on this dataset. MiniGPT4~\cite{zhu2023minigpt} performs stage-2 training using image captions generated by ChatGPT, which are longer than the data used in the first stage. InstructBLIP~\cite{liu2023improved} uses a much wider range of vision-language instruction data, covering both template-based converted data and LLM-generated data. Cheetah~\cite{li2023finetuning} generates synthetic interleaved image-text data and finetunes LVLM on these data to enhance the instruction-following ability of LVLM.
However, this direct application of LVLMs to image reasoning tasks often results in suboptimal performance due to their inherent limitations in capturing the full context and nuanced details present in complex visual scenes.

\paragraph{Multi-modal CoT Reasoning.}
A brute-force attempt to solve zero-shot image reasoning is adopting the recent common practice in machine reasoning, i.e., chain-of-thought (CoT) prompting technique \cite{wei2022chain,Zheng_NeurIPS2023,lu2022learn,zhang2023multimodal}, to zero-shot image reasoning. It employs a step-by-step exploitation, allowing the model to mimic human-like reasoning processes by breaking down complex tasks into simpler thought chains, however unable to harness the LVLM to explore visual nuances for prior knowledge elicitation. 
What's worse, CoT reasoning is usually regarded as an emergent ability \cite{Wei2023emergent, liu2023emergent} when LLMs are scaled up, but an open-source LVLM is usually built upon on 10B-scale LLM (e.g., LLaMA-13B \cite{touvron2023llama}), thus limiting its capacity to fully realize the potential of this technique in the context of diverse and intricate visual reasoning challenges \cite{Fu2023COTHUB}. 
By comparison, in this work we propose QVix as an innovative approach that addresses these challenges, fostering a more holistic and nuanced information of visual data in LVLMs.

\paragraph{GPT-assisted Image Reasoning.}
LLMs such as GPT~\cite{openai2023gpt4} contain massive prior knowledge and strong reasoning ability as they are pre-trained on huge corpora. These methods enable collaboration between LLM and LVLMs to solve vision-language tasks. IdealGPT~\cite{you2023idealgpt} uses an LLM to generate sub-questions, an LVLM to answer the sub-questions, and an LLM to perform the final reasoning. \citeauthor{chen2023see}~\cite{chen2023see} utilize a detection model and an LVLM to generate the caption for each visual concept in the image, an LLM to attend to the key concepts and generate an answer, and finally, an LLM to generate the supporting rationale to the answer and verify the generated rationale. 
However, existing methodologies predominantly engage the LLM primarily for final reasoning~\cite{Zheng_NeurIPS2023,chen2023see,you2023idealgpt,wu2023visual}, relegating the LVLM to a subsidiary role in accomplishing sub-tasks. Consequently, the full reasoning potential of LVLMs remains underutilized in these frameworks. In contrast, our work focuses on unlocking and amplifying the intrinsic reasoning capabilities of LVLMs. By doing so, we aim to more thoroughly harness their power in complex vision-language tasks, ensuring a more integrated and effective utilization of their joint strengths in reasoning and visual comprehension.

\begin{figure*}[htb]
  \centering
   \includegraphics[width=1\linewidth]{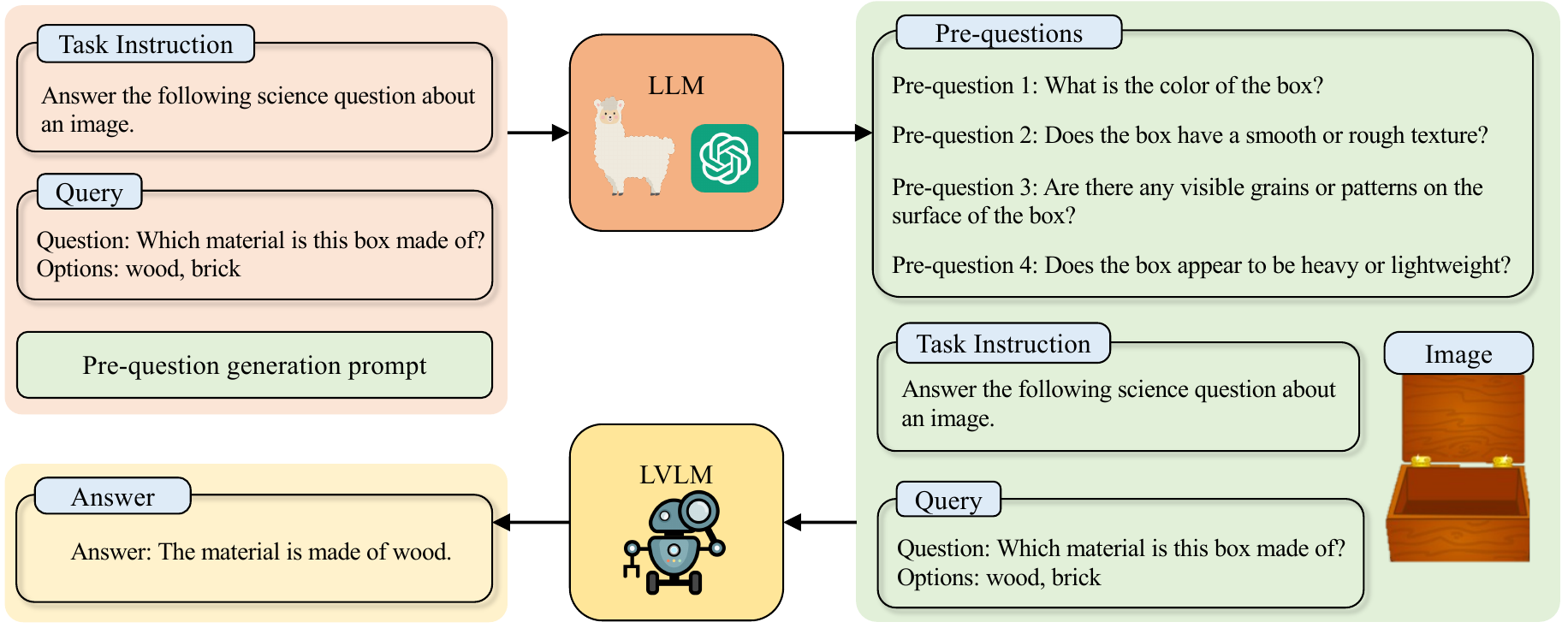}

   \caption{\footnotesize \textbf{Q}uestion-driven \textbf{Vi}sual E\textbf{x}ploration (QVix) prompting for zero-shot image reasoning. Pre-question generation prompt can be found in Fig.~\ref{fig:prompt}.}
   \label{fig:pipeline}
   \vspace{-1.em}
\end{figure*}

\section{Question-Driven Visual Exploration}

We will elaborate on our proposed prompting framework, \textbf{Q}uestion-driven \textbf{Vi}sual E\textbf{x}ploration (QVix), for zero-shot image reasoning tasks.

\subsection{LVLM as Zero-shot Task Solver}

\paragraph{Task Definition.}
In general, given an image $x$ and an associated query\footnote{We use the term `query' to represent the original question in an example, distinguished with the generated `question' in the later literature.} $q$ in a task $t$, zero-shot image reasoning aims to predict an answer $a$ without having seen examples of the specific task $t$ during training. This requires the model to not only analyze the visual content in $x$ but also understand and relate it to the textual query $q$. 
Therefore, we need to find a mapping function $f^t$ from a pair of the image $x\in\gX$ and the query $q\in\gQ$ to an answer $y\in\gY$. That is,
\vspace{-0.5em}
\begin{align}
    f^t: \gX \times \gQ \rightarrow \gY.
\end{align}

This applies to most computer vision or vision language tasks, no matter its output type (i.e., discriminative and generative), category (e.g., fine-grained image classification, visual question answering), dataset (e.g., ImageNet, ScienceQA). 

Traditionally, we need to train a model $\theta^{\text{(t)}}$ over supervised data in the image reasoning task $t$, and perform the inference upon
\vspace{-0.5em}
\begin{align}
    y \sim P(\ry|x, q; \theta^{t}). 
\end{align}

Attributed to recently advanced LVLM, it's possible to define a unified mapping function applicable to a wide range of tasks \cite{xu2023lvlm}, $\forall t\in\gT$, which can be parameterized by $\theta^{\text{(vl)}}$:
\vspace{-0.5em}
\begin{align}
    & f: \gX \times \gT \rightarrow \gY, \\
    & y \sim P(\ry|\compose(t, x, q); \theta^{\text{(vl)}}) \label{equ:vanilla_reason},
\end{align}

where $\compose(\cdot)$ denotes combining the task instruction $t$ with the inputs ($x$, $q$) by using a template in line with prompt heuristics of $\theta^{\text{(vl)}}$. 

Despite the impressive capabilities of LVLMs in handling diverse vision-language tasks, the vision encoder is adapted to the LLM by a coarse-grained alignment, resulting in a primary challenge that LVLMs tend to focus predominantly on the explicit elements of the image that are directly referenced in the text prompt, i.e., a form of tunnel vision. 
This tunnel vision narrows the scope of visual exploration, causing LVLMs to overlook subtler but equally important visual cues and context. 

To alleviate this problem, we propose QVix composed of two sequential prompting processes, i.e., good question exploring (\S\ref{sec:goodquest}) and question-conditioned reasoning (\S\ref{sec:questcond}).

\subsection{Good Pre-Question Exploration} \label{sec:goodquest}
This good pre-question exploring process aims to systematically expand the observational scope of LVLMs beyond the initially presented visual-textual context. By generating a set of contextual task-aware pre-questions by a model, this approach intends to direct the LVLM's attention toward previously unexplored or underexplored areas of the visual input. This is based on the premise that the divergent questions can drive the LVLM to explore broad regions of an image and uncover deeper layers of meaning and information in the image, which would otherwise remain untapped using conventional methods.

To achieve this, we leverage a generative model to produce a set of pre-questions. 
We generate the pre-questions based on the task $t$ and the input query $q$) regardless of the image $x$ for two reasons: i) without the constraints of the input image $x$, it can improve the openness of the resulting questions and thus increase the diversity of our visual exploration, and ii) without the modality of images, all other inputs (i.e., $t$ and $q$), as well as the output, is in natural language, expanding model options, either open- or close-source, either pure LLM or LVLM.
As such, we can define
\begin{align}
    \sQ_t^{\text{(gd)}} \!=\! \{q_t^{\text{(gd)}}| q_t^{\text{(gd)}} \!\sim\! P(\rq_t^{\text{(gd)}}|\genprompt(t,q);\theta^{\text{(gen)}}\}_1^M,
\end{align}
where $M$ denotes the number of generated `good' exploratory pre-questions, and $\genprompt(t,q)$ denotes a prompting template to generate exploratory pre-questions for the given task instruction $t$ and the text original query $q$. Please refer to the prompt in Fig.~\ref{fig:prompt} for an implementation of $\genprompt(\cdot)$. 

\begin{figure}
\begin{center}
\fcolorbox{black}{gray!10}{\parbox{.9\linewidth}{I need to answer the following main question about an image: 
\\ \hspace*{\fill} \\
\{Task Instruction\}
\\ \hspace*{\fill} \\
\{Query\}
\\ \hspace*{\fill} \\
Your goal is to design 4 pre-questions. Pre-questions should focus on important contextual information in the image useful for answering the main question.
\\ \hspace*{\fill} \\
Here are the rules you should follow when listing the pre-questions:

Each pre-question should be short and easy to understand.

Pre-questions should focus on context visual clues of the image.

Pre-questions should provide clues to answer the main question.
\\ \hspace*{\fill} \\
Format Example:

Pre-question 1: xxxx

Pre-question 2: xxxx 

Pre-question 3: xxxx

Pre-question 4: xxxx
}}
\end{center}
   \vspace{-1.em}
 \caption{\footnotesize Pre-question generation prompt in QVix.}
 \label{fig:prompt}
 \vspace{-2.em}
\end{figure}

Note that i) although this process is based on a model $\theta^{\text{(gen)}}$ operating on pure texts, the model can be an arbitrary generative model, no matter an LLM $\theta^{\text{(llm)}}$ or LVLM $\theta^{\text{(vl)}}$, and ii) again, `good' doesn't mean the asked questions are optimal but just good to provide extra prompts for the downstream LVLM to guide the LVLM towards a more thorough examination of the visual data.

\paragraph{Remark.} The inspiration for this approach comes from the way expert human analysts approach visual data: by asking probing, insightful questions to uncover hidden details and meanings. This motivates us to harness the natural language understanding capabilities of LLMs to formulate these questions, thereby enabling LVLMs to break free from the confines of their initial visual-textual alignment and embark on a  wide-ranging visual exploration.

\subsection{Question-conditioned Reasoning} \label{sec:questcond}

The aim of question-conditioned reasoning is to integrate the additional context and insights gained from the Good Pre-Question Exploring phase into the LVLM's reasoning process. This approach seeks to utilize the newly generated pre-questions as a means to refine, redirect, or expand the LVLM's interpretive focus. The intuition behind this is that the LVLM, equipped with new pre-questions and perspectives, can reassess and reinterpret the visual data in a more nuanced and informed manner. This method mirrors the dynamic and adaptive nature of human reasoning, where new information leads to the reevaluation and enrichment of existing understanding.

To implement this, we simply inject $\sQ_t^{\text{(gd)}}$ into reasoning formulation shown in Eq.(\ref{equ:vanilla_reason}), i.e.,
\begin{align}
    y \sim P(\ry|\conditionprompt(t, x, q, \sQ_t^{\text{(gd)}}); \theta^{\text{(vl)}}). \label{equ:final_reason}
\end{align}
Here, $\conditionprompt(\cdots)$ denotes a simple prompt template for this question-conditioned reasoning, where we directly concatenate image $x$, pre-questions  $\sQ_t^{\text{(gd)}}$, task instruction $t$ and query $q$ as shown in Fig.~\ref{fig:pipeline}.

\section{Experiments}

\begin{table*}[t]
\footnotesize
\centering
    \begin{tabular}{lcccccc}
    \toprule
    \multirow{2}{*}{Method}&\multicolumn{3}{c}{Subject}&\multicolumn{2}{c}{Grade} & \multirow{2}{*}{Average} \\
    \cmidrule(r){2-4} \cmidrule(r){5-6} 
    ~&NAT&SOC&LAN&G1-6&G7-12&~\\
    \midrule
    LLaVA &37.0&61.5 &33.3 &52.3 &30.5 &46.2 \\
    MiniGPT &45.2 &51.5 &\underline{38.1} &50.6 & 39.1 &47.4 \\
    InstructBLIP&43.9&58.1&\textbf{47.6}&53.1&\underline{39.4}&49.3\\  
     \midrule
    QVix (Self-generated questions)&\underline{47.8}&\underline{62.9}&\textbf{47.6}&\underline{58.9}&\underline{39.4}&\underline{53.5}\\
    QVix (GPT-generated questions)&\textbf{48.0}&\textbf{67.1}&\underline{38.1}&\textbf{60.6}&\textbf{40.5}&\textbf{55.0}\\
    \bottomrule
    \end{tabular}
    \caption{\footnotesize Zero-shot \textbf{VQA} accuracy (\%) on ScienceQA. Question categories: NAT = natural science, SOC = social science, LAN = language
science, G1-6 = grades 1-6, G7-12 =
grades 7-12. QVix in the table is applied to InstructBLIP. The \textbf{first} and \underline{second}-best methods for each category are highlighted.\looseness-1}
    \label{tab:scienceqa}
       \vspace{-1.5em}
\end{table*}

\begin{table}[htb]
\footnotesize
	\centering
\setlength\tabcolsep{3pt}
    \begin{tabular}{lccc}
    \toprule
    Method&Flowers102&Oxford-IIIT Pet&FGVCAircraft\\
    \midrule
    LLaVA & 36.1 & 39.9 & 15.5 \\
    MiniGPT & 51.5 &63.9 & 14.8\\
    InstructBLIP&59.5&69.8&25.1\\
         \midrule
    QVix (Self-generated) & \underline{66.9} &\underline{76.1} & \underline{26.8}\\
    QVix (GPT-generated)&\textbf{67.8}&\textbf{80.5}&\textbf{27.7}\\
    \bottomrule
    \end{tabular}
         \caption{\footnotesize Zero-shot \textbf{image classification} accuracy (\%) on Fine-grained Image Classification. QVix is applied to InstructBLIP.}
         \label{tab:imgcls}
            \vspace{-1em}
\end{table}

In this section, we evaluate the proposed QVix in three vision-language tasks: ScienceQA, fine-grained image classification, and Visual Entailment. We show that QVix achieves superior zero-shot performance on these tasks generally compared to other existing baselines. We apply QVix to different LVLMs to show its generality. We conduct a case study to illustrate how QVix attends to important visual areas. Finally, We conduct a thorough analysis of key components of QVix.
\vspace{-1em}
\paragraph{Datasets.} We evaluate our proposal on 5 datasets from three categories of reasoning
tasks: Visual Question Answering, fine-grained image classification and Visual Entailment. For Visual Question Answering, we use ScienceQA~\cite{lu2022learn} which represents a multimodal benchmark encompassing multiple-choice questions across a wide range of scientific subjects. For our evaluation purposes, we exclusively utilized those samples from the test set that are accompanied by images, obtaining a total of 2017 image-question pairs. For fine-grained image classification, we use three popular datasets: Flowers102~\cite{nilsback2008automated}, Oxford-IIIT Pet~\cite{parkhi2012cats} and FGVCAircraft~\cite{maji13fine-grained} which require detailed differentiation of the species of flowers, breeds of pets, and models of cars respectively. For Visual Entailment, we use SNLI-VE which is derived from the Stanford Natural Language Inference (SNLI) initiative by Bowman~\cite{bowman2015large}, which is fundamentally a text entailment (TE) task built upon Flicker30k~\cite{plummer2015flickr30k} data. 
In our study, for each dataset, a random selection of 1000 samples was made from the validation/test split. We then assessed our method in a zero-shot context, using accuracy as the metric for evaluation.

\begin{figure}[t]
\vspace{-1em}
  \centering
   \includegraphics[width=0.9\linewidth]{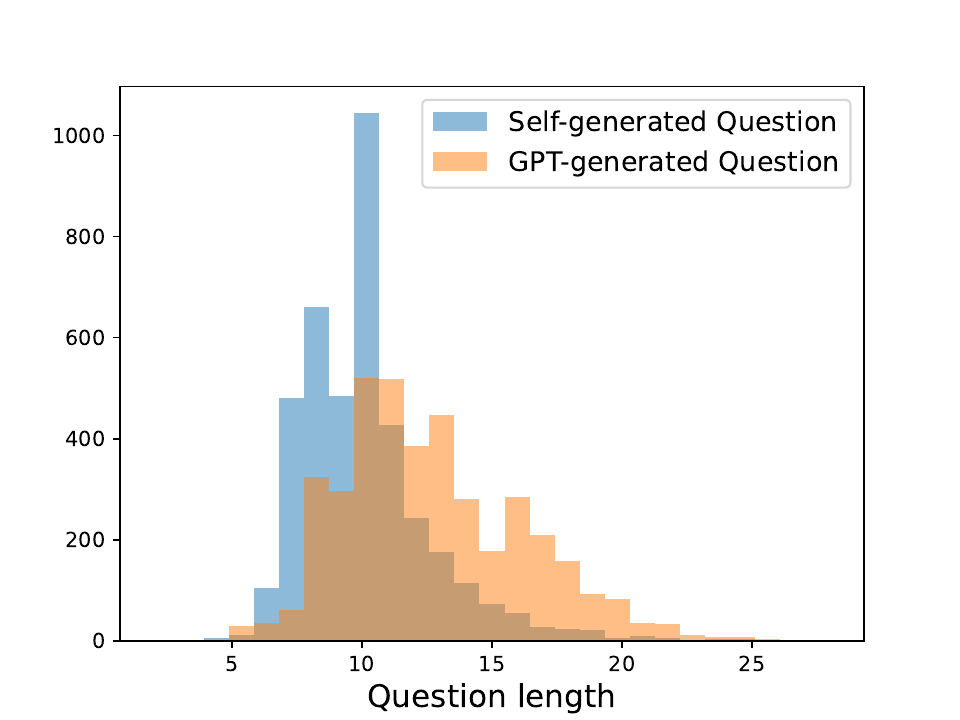}
   \vspace{-1em}
   \caption{\footnotesize Histograms of the length of the self(VLM)-generated and GPT-generated pre-questions using the prompt in Fig.~\ref{fig:prompt}.}
   \label{fig:hist}
      \vspace{-1.em}
\end{figure}
\vspace{-1em}

\paragraph{Models.} 
We employ two strategies for generating pre-questions. Initially, we tap into GPT~\cite{zhu2023chatgpt} for its advanced reasoning and extensive knowledge, utilizing the 'GPT-3.5-turbo' API for question generation, denoted as ``\textit{GPT-generated questions}''. However, due to the high cost and speed limit of the GPT API, which is impractical for large-scale use, we alternatively generate pre-questions using the embedded LLM in open-source VLMs, greatly increasing our method's usability, denoted as ``\textit{self-generated questions}''. 
For compared baselines, we try three pretrained LVLMs, LLaVA~\cite{liu2023visual}, MiniGPT~\cite{zhu2023minigpt}, and InstructBLIP to do zero-shot reasoning.
\subsection{ScienceQA}

ScienceQA aims to evaluate the model's capacity to navigate and interpret complex scientific information. Each instance in this dataset encompasses a meticulously crafted scientific question, spanning diverse domains including natural science, social science, and language science. 
The reasoning process of the model may encompass a range of complexities: from straightforward factual queries to intricate problems, understanding of scientific processes, and application of theoretical concepts. 
The performance of various methods on the ScienceQA dataset is examined in Table~\ref{tab:scienceqa}. From Table~\ref{tab:scienceqa}, it's evident that the InstructBLIP method outperforms both LLaVA and MiniGPT in every subject and grade category. Notably, the proposed QVix, whether using self-generated questions or GPT-generated questions,  demonstrates substantial improvement over baselines, particularly in the NAT and SOC subjects, reflecting its strength in generating context-relevant queries. With self-generated questions, QVix achieves a 53.5\% average accuracy, and outperforms the base model InstructBLIP by 4.2\%. When equipped with GPT-generated questions, QVix achieves the highest overall average score of 55.0\%, outperforming that using self-generated questions, which demonstrates that the richer prior knowledge of a more powerful language model can provide better pre-questions.

\begin{figure*}[htb]
    \footnotesize
  \centering
   \includegraphics[width=1\linewidth]{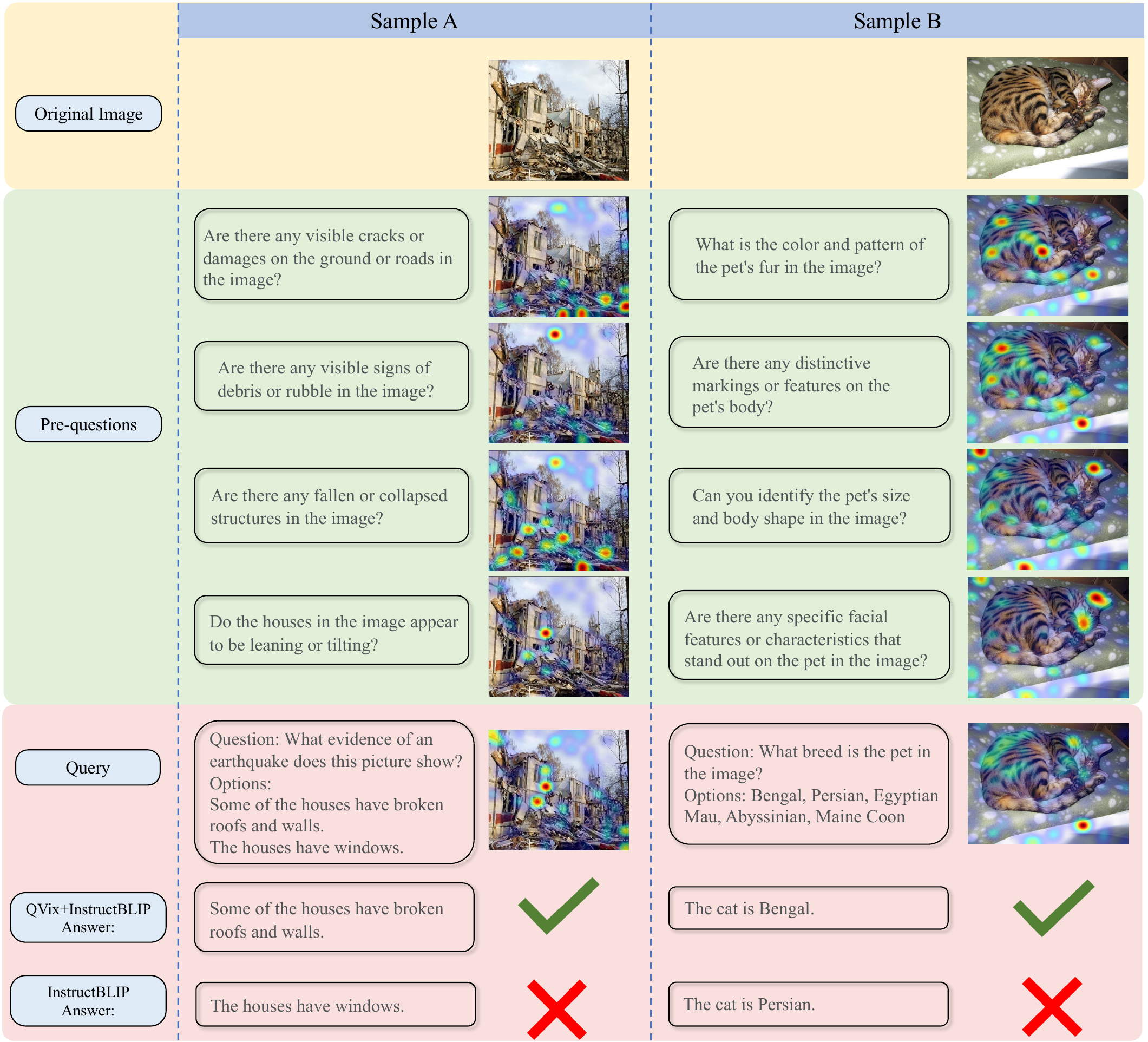}

   \caption{\footnotesize Case studies of QVix when applied to VQA tasks. The question-image attention maps are generated by the Q-former of InstructBLIP.}
   \label{fig:case}
   \vspace{-1em}
\end{figure*}

\subsection{Fine-grained Image Classification}

\begin{table}[t]
    \footnotesize
    \centering
    \begin{tabular}{lc}
    \toprule
    Method&Accuracy (\%)\\
    \midrule
    MiniGPT & 35.1 \\
    LLaVA & 40.3  \\
    InstructBLIP&34.5\\
    \midrule
    QVix (Self-generated questions) & \underline{43.0} \\
    QVix (GPT-generated questions)&\textbf{50.1}\\
    \bottomrule
    \end{tabular}
    \caption{\footnotesize Zero-shot accuracy (\%) on \textbf{Visual Entailment}. QVix in the table is applied to InstructBLIP.}
    \label{tab:ve}
    \vspace{-2em}
\end{table}


Fine-grained Image Classification is a sophisticated task, where the objective is to differentiate between highly similar sub-categories within a broader category. 
Previous LVLMs performed  poorly in fine-grained classification~\cite{xu2023lvlm}. 
Table~\ref{tab:imgcls} presents a comparative analysis of different computational methods applied to fine-grained image classification tasks within three specific datasets: Flowers101, Oxford-IIIT Pet, and FGVC-Aircraft. The table highlights the superior performance of our method, QVix, especially when utilizing GPT-generated questions. QVix with GPT-generated questions outshines other methods by achieving the highest accuracy scores across all datasets: 67.8\% for Flowers101, 80.5\% for Oxford-IIIT Pet, and 27.7\% for FGVC-Aircraft. This performance underscores the effectiveness of QVix's strategy of integrating advanced language models to understand and classify images with high granularity.
Comparatively, QVix with self-generated questions also performs robustly, coming in a close second to its GPT-enhanced counterpart, and still outperforms the other listed methods in all datasets. The result indicates that our self-generated questions method is highly competitive, with only a marginal difference when compared to the GPT-generated questions, suggesting that the underlying architecture of QVix is solid even without GPT's assistance.
The success of QVix can be attributed to its ability to generate more contextually relevant  pre-questions that can guide the model to focus on more detail-oriented information within the image that is beneficial for fine-grained classification.

\subsection{Visual Entailment}

The task of Visual Entailment involves prompting the model to ascertain if the text is semantically entailed by the image. Each instance contains a coupled set composed of an image and a corresponding text hypothesis, accompanied by three potential answers: entailment, neutral, and contradiction. 
Addressing this tri-category classification challenge in a zero-shot framework is notably arduous.
 Table ~\ref{tab:ve} presents a striking demonstration of the effectiveness of our QVix method in the task of Visual Entailment. Significantly, the GPT-generated question variant of QVix achieves an accuracy of 50.1\%, surpassing all other methods by a notable margin. This highlights the advanced understanding and synthesis capabilities of the GPT model when it comes to generating contextually relevant and complex questions for improving visual data understanding, a key component in visual entailment.
Equally important is the performance of QVix with self-generated questions, which records a 43.0\% accuracy. This outstrips the MiniGPT, LLaVA, and InstructBLIP methods by 7.9\%, 2.7\%, and 8.5\%, respectively.
The success of QVix suggests that the questions with contextual richness and diversity can guide the VLM to comprehensively understand the semantic information of the entire image based on context and details, which is a key demand for Visual Entailment.

\begin{figure*}
    \footnotesize
  \centering
   \includegraphics[width=1\linewidth]{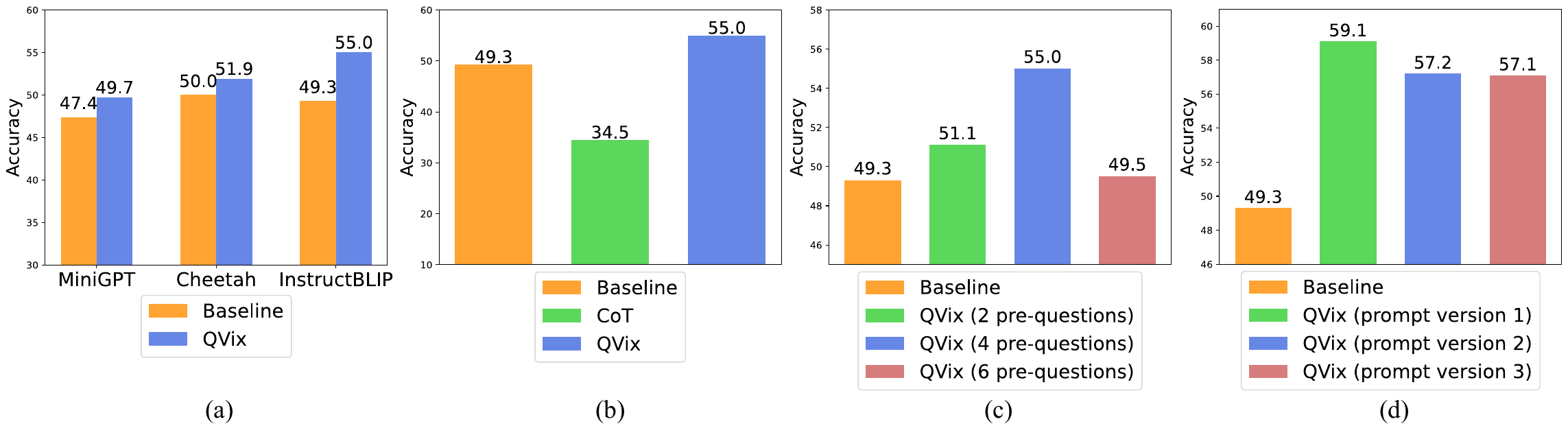}
      \vspace{-2em}
	\caption{\footnotesize Analysis of QVix on ScienceQA. (a) QVix improves different LVLMs' zero-shot reasoning performance. (b) QVix vs. zero-shot CoT. (c) QVix with different numbers of pre-questions. (d) QVix with GPT4 re-written prompts (Fig.~\ref{fig:gpt re1}-\ref{fig:gpt re3}) paraphrased from the original prompt in Fig.~\ref{fig:prompt}. ``Baseline'' refers to zero-shot reasoning without QVix or CoT. InstructBLIP is the LVLM for (b)-(d).}
   \vspace{-1em}
	\label{fig: analysis}
\end{figure*}

\subsection{Analysis}
\label{sec:analysis}
In this subsection, we conduct a thorough analysis of QVix. We show the generality and stability of QVix, perform a case study, and compare QVix with CoT.
In all experiments, we use the test split of ScienceQA, which in our findings is enough to distinguish different model choices.

\vspace{-1em}

\paragraph{Generality of QVix.}
QVix is a general prompting framework that can be applied to different LVLMs. Here we mainly try three different LVLMs: MiniGPT, Cheetah, and InstructBLIP. MiniGPT and InstructBLIP are two highly popular LVLMs, while cheetah is the latest LVLM trained on image-text interleaved data which has strong instruction-following ability. The results are shown in Fig.~\ref{fig: analysis}(a). 
Across the board, the implementation of QVix results in a consistent improvement in accuracy over the baseline configurations for each of the LVLMs tested. Cheetah, as a state-of-the-art LVLM, demonstrates its robust instruction-following capabilities with a baseline accuracy of 50.0\%. With the integration of QVix, there is a further gain of 1.9\% in accuracy. This increment, while modest, is indicative of QVix's potential to augment even the latest LVLMs that are already optimized for processing complex multimodal data. 
InstructBLIP, another popular LVLM, shows the most significant improvement with QVix, jumping from a baseline accuracy of 49.3\% to 55.0\%.  QVix can serve as a valuable addition to the LVLMs, offering a versatile approach to boosting model performance across a variety of architectures and training paradigms.

\paragraph{Case Study.} 
To understand and explain how QVix explores visual information and makes the final reasoning, we provide a case study for baseline and QVix in Fig.~\ref{fig:case}. We visualize the cross-attention map between text and image. Compared to the baseline, QVix can attend to important visual areas through those pre-questions. In Sample A, by asking ``\textit{Are there any fallen or collapsed structures in the image?}'', the model successfully attends to ``\textit{broken roofs and walls}'' on the ground, which leads to the correct final reasoning. While the baseline only attends to irrelevant areas such as ``\textit{windows}''. Similarly, Sample B's cross-attention maps show QVix's attention to distinctive features of the cat, e.g., ``\textit{eyes}'' and ``\textit{mouse}'', correctly recognizing its breed as Bengal, which is reflective of QVix's ability to parse finer details and patterns in the image. These cases reveal how QVix prompts the model to scrutinize specific visual details that are crucial for accurate image reasoning. This contrasts sharply with the baseline approach, which lacks this directed exploratory mechanism, resulting in attention being cast toward areas that do not contribute to solving the task at hand.

\paragraph{Comparison with Zero-Shot CoT.}
Here we compare QVix with zero-shot CoT~\cite{wei2022chain}. For zero-shot CoT, we use the prompt \emph{``Let's think step by step by looking at the image.''} to generate rationale and then condition on the rationale to perform the final reasoning. We find that LVLM fails to generate reasonable rationales as the CoT reasoning is usually regarded as an emergent ability but the open-source LVLM is usually built upon 10B-scale LLM which lacks CoT reasoning ability. As indicated in Fig.~\ref{fig: analysis}(b), CoT's performance falls short, not only relative to QVix but also when compared to the baseline. In contrast, QVix actively steers the LVLM to scrutinize critical visual regions, thereby significantly bolstering the reasoning efficacy of open-source LVLMs.

\paragraph{Number of Pre-Questions.}
A key parameter in QVix is the number of pre-questions. Too few pre-questions may fail to provide sufficient contextual information, while an excess of pre-questions could render the model's reasoning overly challenging. 
We study the proper number of pre-questions in Fig.~\ref{fig: analysis}(c). 
When two pre-questions are used, there is a modest increase in accuracy to 51.1\%, suggesting that even a minimal addition of context can be beneficial.
However, the most notable improvement is observed when the number of pre-questions is increased to four, with accuracy peaking at 55.0\%. This finding implies that four pre-questions strike a delicate balance between providing adequate context for the model to process the task effectively and avoiding an information overload that could potentially confuse the model.
Interestingly, when the number of pre-questions is further increased to six, accuracy drops to 49.5\%, which is marginally higher than the baseline but lower than with two pre-questions.  This decrease may indicate that additional context does not translate into better performance but may hinder the model by introducing unnecessary complexity.

\paragraph{Prompt Stability Analysis.}
In order to analyze the stability of our hand-written pre-question generation prompt, we employ GPT to rewrite our manually designed prompts in three different versions. The results are shown in Fig.~\ref{fig: analysis}(d). This analysis indicates that while there may be slight fluctuations in outcomes with different prompt versions,
the QVix method contributes to stable and enhanced performance compared to the baseline. The ability of QVix to consistently outperform the baseline through various rewrites confirms its potential as a reliable and effective tool for improving the accuracy of vision-language tasks.

\section{Conclusion}
This paper proposes QVix, a novel framework for enhancing LVLMs' zero-shot image reasoning by countering the ``\textit{tunnel vision}" effect through a question-driven visual exploration strategy. Our approach leverages the inherent natural language understanding and prior knowledge of LLMs to generate input-exploratory pre-questions, guiding LVLMs to explore visual content more comprehensively and uncover subtle or peripheral details. Empirical evaluations demonstrate that QVix markedly improves the performance of various LVLMs on diverse vision-language tasks, including ScienceQA, fine-grained image classification, and Visual Entailment.

{
    \small
    \bibliographystyle{ieeenat_fullname}
    \bibliography{arxiv}
}
\clearpage
\setcounter{page}{1}
\maketitlesupplementary
\begin{table*}[htb]
    \footnotesize
    \centering
    \begin{tabular}{p{70pt}p{250pt}p{70pt}}
    \toprule
    Dataset Name&Dataset Description&Evaluation Data \\
    \midrule
    ScienceQA & ScienceQA~\cite{lu2022learn} is a multimodal benchmark containing multi-choice questions with
a diverse set of science topics. In our evaluation, we only use the samples with images
in the test set. & 1000 (test) \\
    \midrule
    Flowers102 & The Oxford 102 Flower dataset~\cite{nilsback2008automated} includes 120 flower categories (Flowers102 for
short) with 40 to 258 images for each class and 8189 images in total, namely 10 images
per class for both train and val and the rest for a test. &1000 (test)\\
   \midrule
    Oxford-IIIT Pet  & The Oxford-IIIT Pet dataset~\cite{parkhi2012cats} comprises 37 categories  with 25
dog breeds and 12 cat ones and 200 images per class. There are 7349 images in total, 3680 train/val images, and 3669 test images. &1000 (test)\\
   \midrule
    FGVC Aircraft & The FGVC-Aircraft dataset contains 10,200 images of aircraft, with 100 images for each of 102 different aircraft model variants, most of which are airplanes.  &1000 (test)\\
    \midrule
    SNLI-VE & SNLI-VE~\cite{bowman2015large} extends the text entailment (TE) task into the visual domain and asks
the model whether the image is semantically entailed, neutral, or contradicted to the
next hypothesis. It is a three-category classification task based on Flicker30k~\cite{plummer2015flickr30k}. &1000 (dev)\\
    \bottomrule
    \end{tabular}
    \caption{\footnotesize Details of the used datasets.}
    \label{tab:data}
\end{table*}

\begin{table*}[htb]
    \footnotesize
    \centering
    \begin{tabular}{p{60pt}p{160pt}}
    \toprule
    Dataset&Task Instruction \\
    \midrule
    ScienceQA & / \\
  \midrule
    Flowers102 & \textit{What breed is the flower in the image?}\\
  \midrule
    Oxford-IIIT Pet  & \textit{What breed is the per in the image?} \\
  \midrule
    FGVCAircraft & \textit{What is the aircraft in the image?}\\
  \midrule
    SNLI-VE& \textit{Predict whether the image semantically entails the textual hypothesis, choose the answer from entailment, neutral, contradiction.} \\
    \bottomrule
    \end{tabular}
    \caption{\footnotesize Detailed task instruction for each dataset.}
    \label{tab:task ins}
\end{table*}

\begin{table*}[htb]
    \footnotesize
    \centering
    \begin{tabular}{p{70pt}p{350pt}}
    \toprule
    Dataset &Pre-questions\\
    \midrule
    ScienceQA & Example 1: 
     What is the shape and color of the marker indicating a city on the map? Are there any labels or names written near the marker on the map?\ Are there any geographical features or landmarks near the marked city on the map? Is the marked city located on the coast or inland on the map?

     Example 2: What is the color of the water in the image? Are there any visible coral reefs in the image? Are there any fish or other marine organisms visible in the image? Is there any evidence of saltwater in the image, such as waves or sea spray?
    \\
    \midrule
    Flowers102 & What colors are present in the flower? Are there any distinctive markings or patterns on the flower? What is the shape and size of the flower compared to other elements in the image? Are there any similar flowers in the image that could provide clues to the breed of the flower? \\
   \midrule
    Oxford-IIIT Pet  & What is the color and pattern of the pet's fur in the image? Are there any distinctive markings or features on the pet's body? Can you identify the pet's size and body shape in the image? Are there any specific facial features or characteristics that stand out on the pet in the image? \\
   \midrule
    FGVC Aircraft & What is the color of the car in the image? Are there any visible logos or brand names on the car? Is the car a sedan or an SUV? Are there any distinguishing features of the car, such as unique headlights or a special design element?  \\
    \midrule
    SNLI-VE & What objects or elements are prominently featured in the image? Are there any interactions or relationships between the objects or elements in the image? What is the overall mood or atmosphere conveyed by the image? Are there any specific visual cues, such as colors, lighting, or expressions on faces, that may indicate the emotional tone of the image?  \\
    \bottomrule
    \end{tabular}
    \caption{\footnotesize The GPT-generated pre-questions for each dataset.}
    \label{tab:pre-question}
\end{table*}

\begin{table*}[t]
\footnotesize
\centering
    \begin{tabular}{lcccccc}
    \toprule
    \multirow{2}{*}{Method}&\multicolumn{3}{c}{Subject}&\multicolumn{2}{c}{Grade} & \multirow{2}{*}{Average} \\
    \cmidrule(r){2-4} \cmidrule(r){5-6} 
    ~&NAT&SOC&LAN&G1-6&G7-12&~\\
    \midrule
    QVix (GPT-generated question)&49.56$\pm$0.51&66.4$\pm$0.45&42.86$\pm$3.89&60.98$\pm$0.65&42.29$\pm$0.29&55.8$\pm$0.54\\
    \bottomrule
    \end{tabular}
    \caption{\footnotesize Randomness analysis of QVix (\%) on ScienceQA. Error bars (mean and std) are
computed over three random trails. QVix is applied to InstructBLIP.}
    \label{tab:random}
       \vspace{-1.5em}
\end{table*}

\section{More Implementation Details}
\label{sec:exp detail}
\subsection{Dataset Details}
We evaluate QVix on 5 datasets from 3 tasks. For each dataset, we randomly draw 1000 samples from its validation/test set for evaluation. The details of these datasets are summarized in Table~\ref{tab:data}.

\subsection{Model Details}
We evaluated four LVLMs: LLaVA~\cite{liu2023visual}, MiniGPT~\cite{zhu2023minigpt}, InstructBLIP~\cite{instructblip} and Cheetah~\cite{li2023finetuning}. For LLaVA, we employ the model using MPT-7b as LLM.
In MiniGPT and Cheetah, we employ Llama2-7b~\cite{touvron2023llama} as their LLM. For InstructBLIP, we employ Vicuna-7b~\cite{vicuna2023} as its LLM. Our choice of InstructBLIP, LLaVA, and MiniGPT is based on their excellent overall performance~\cite{xu2023lvlm} and their high popularity. We choose Cheetah due to its state-of-the-art instruction-following ability.

\subsection{Prompts used in QVix}
We generate pre-questions using the prompt template illustrated in Fig.~\ref{fig:prompt}, where we need to incorporate the task instruction and query into this template. For ScienceQA, we generate pre-questions for each instance since the query for each instance is unique in ScienceQA. For the other four datasets,  we generate a set of pre-questions for the whole dataset since the query for different instances in these datasets is the same. The detailed task instruction is shown in Table~\ref{tab:task ins}. For ScienceQA, the task instruction is empty and we use the query for each instance to generate pre-questions. For the other four datasets, we use the task instruction in Table~\ref{tab:task ins} to generate pre-questions. The GPT-generated pre-questions for each dataset are given in Table~\ref{tab:pre-question}.

\subsection{Fine-grained Image Classification}

Previous LVLMs performed poorly in fine-grained classification~\cite{xu2023lvlm}. 
Owing to the large number of categories, typically in the hundreds, it becomes challenging for LVLM to process when all categories are presented as input options. Therefore, we first utilize CLIP model to select the top 5 categories and ask LVLM to predict the class from the 5 choices.  

\begin{figure*}[htb]
    \footnotesize
  \centering
   \includegraphics[width=1\linewidth]{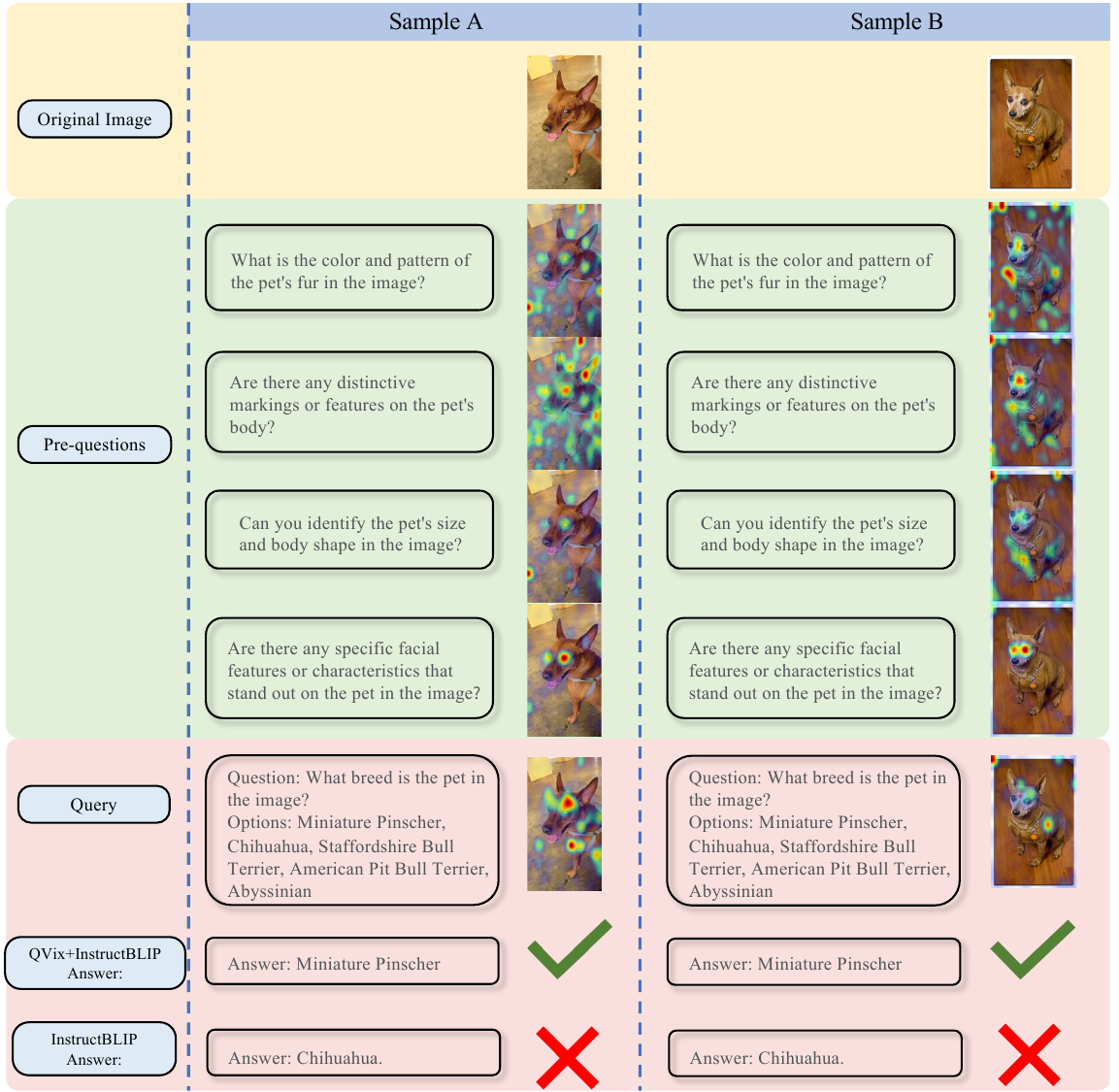}

   \caption{\footnotesize Case study of QVix better extracting detailed information. The attention map is generated by the Q-former of InstructBLIP. }
   \label{fig:case detail}
   \vspace{-1em}
\end{figure*}

\begin{figure*}[htb]
    \footnotesize
  \centering
   \includegraphics[width=1\linewidth]{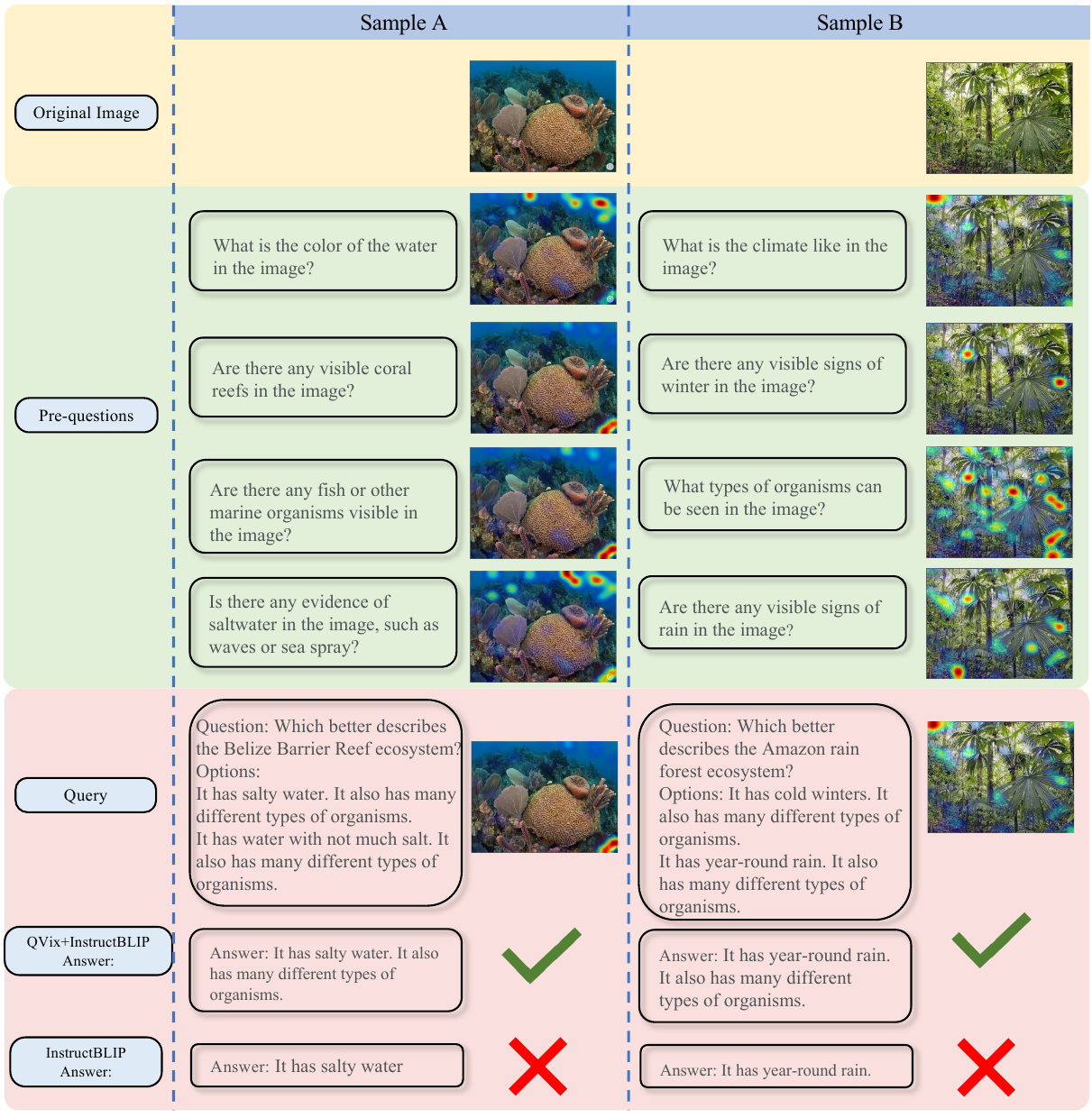}

   \caption{\footnotesize Case study of QVix better extracting contextual information. The attention map is generated by the Q-former of InstructBLIP. }
   \label{fig:case context}
   \vspace{-1em}
\end{figure*}

\begin{figure*}[htb]
    \footnotesize
  \centering
   \includegraphics[width=0.8\linewidth]{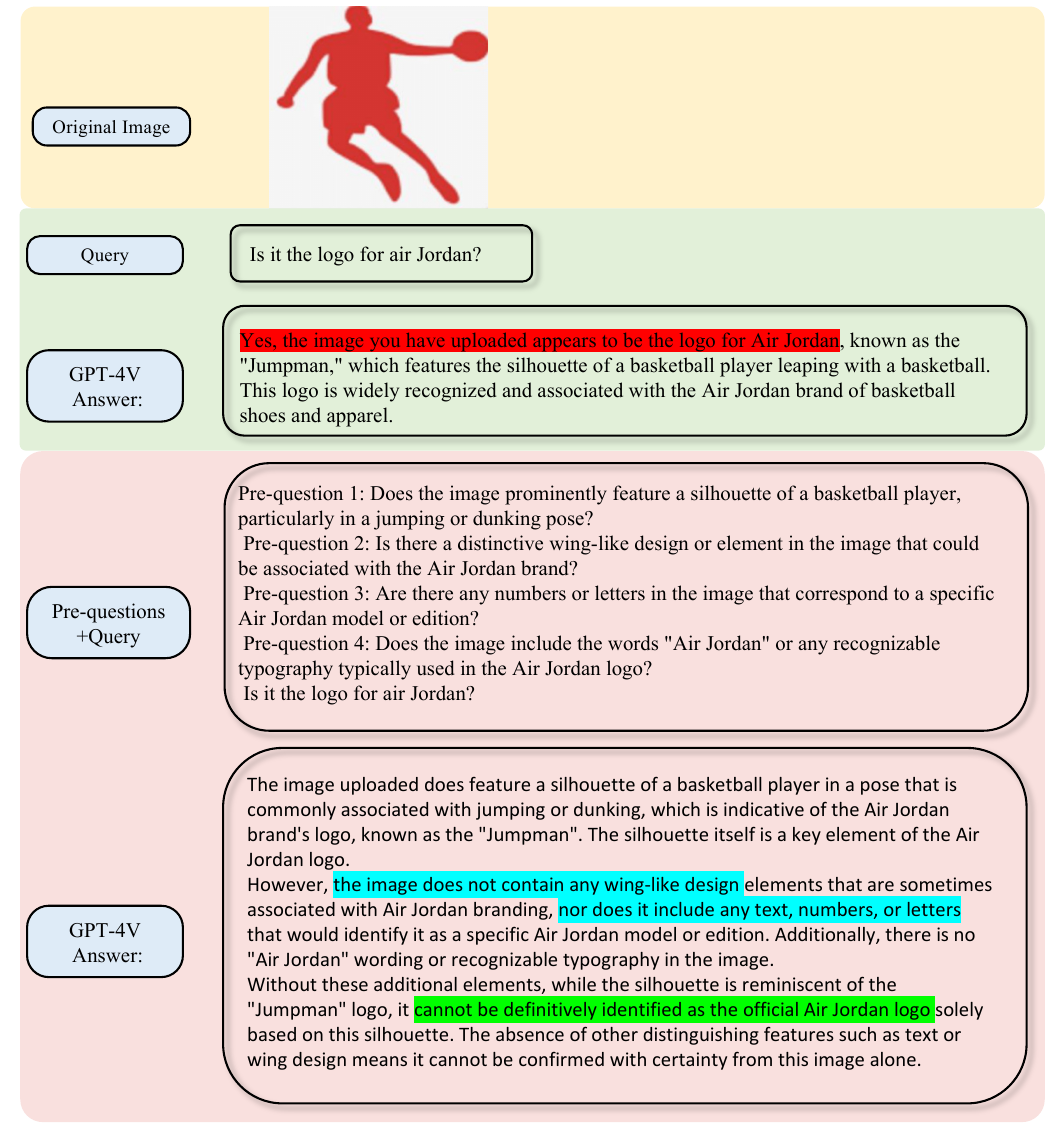}

   \caption{\footnotesize Case study of QVix applied to GPT-4V. We highlight the wrong answer (red), correct answer (green), and rationale (blue) leading to the correct answer. }
   \label{fig:gpt 1}
   \vspace{-1em}
\end{figure*}

\begin{figure*}[htb]
    \footnotesize
  \centering
   \includegraphics[width=0.8\linewidth]{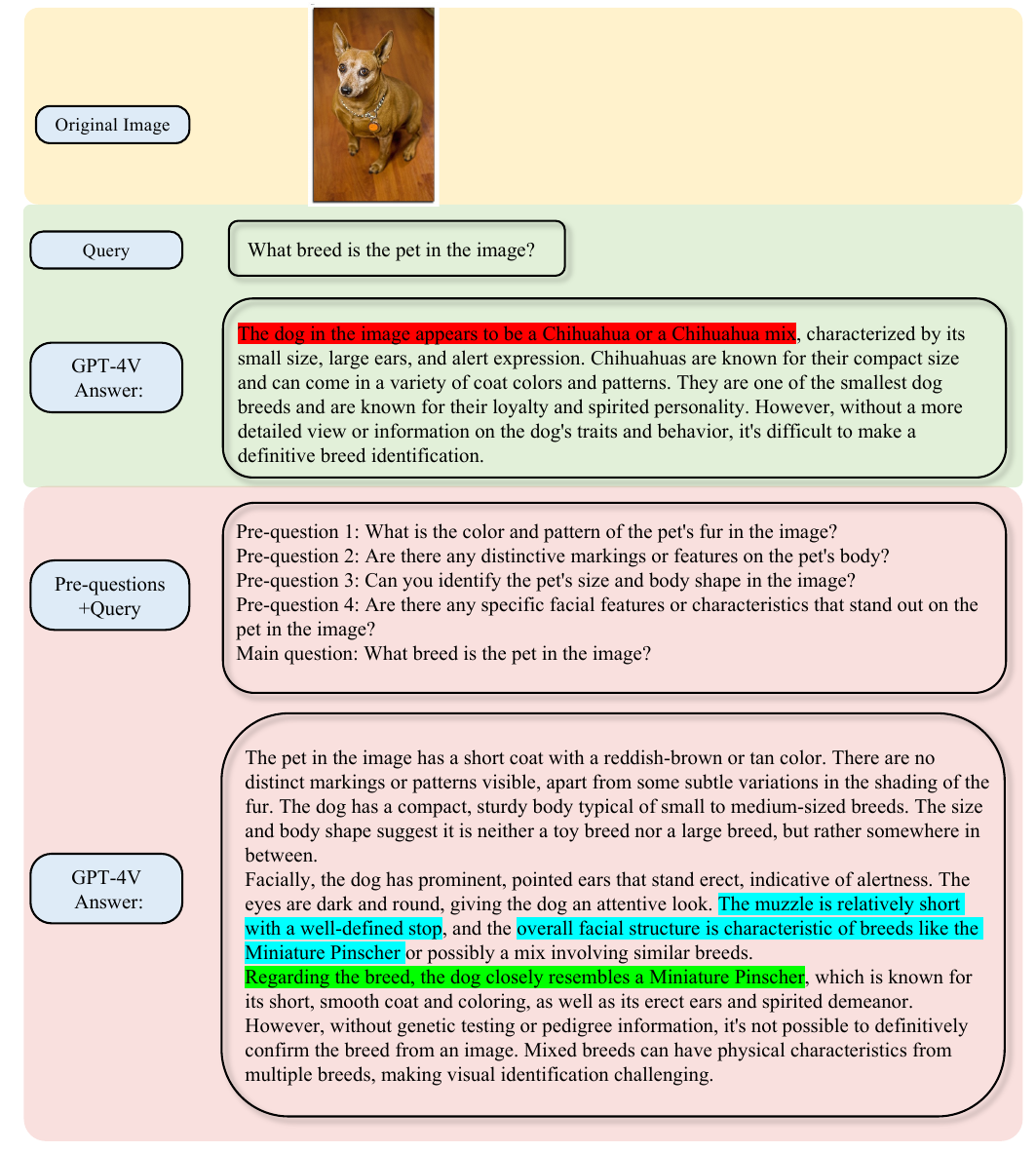}

   \caption{\footnotesize Case study of QVix applied to GPT-4V. We highlight the wrong answer (red), correct answer (green), and rationale (blue) leading to the correct answer. }
   \label{fig:gpt 2}
   \vspace{-1em}
\end{figure*}

\begin{figure*}[htb]
    \footnotesize
  \centering
   \includegraphics[width=0.55\linewidth]{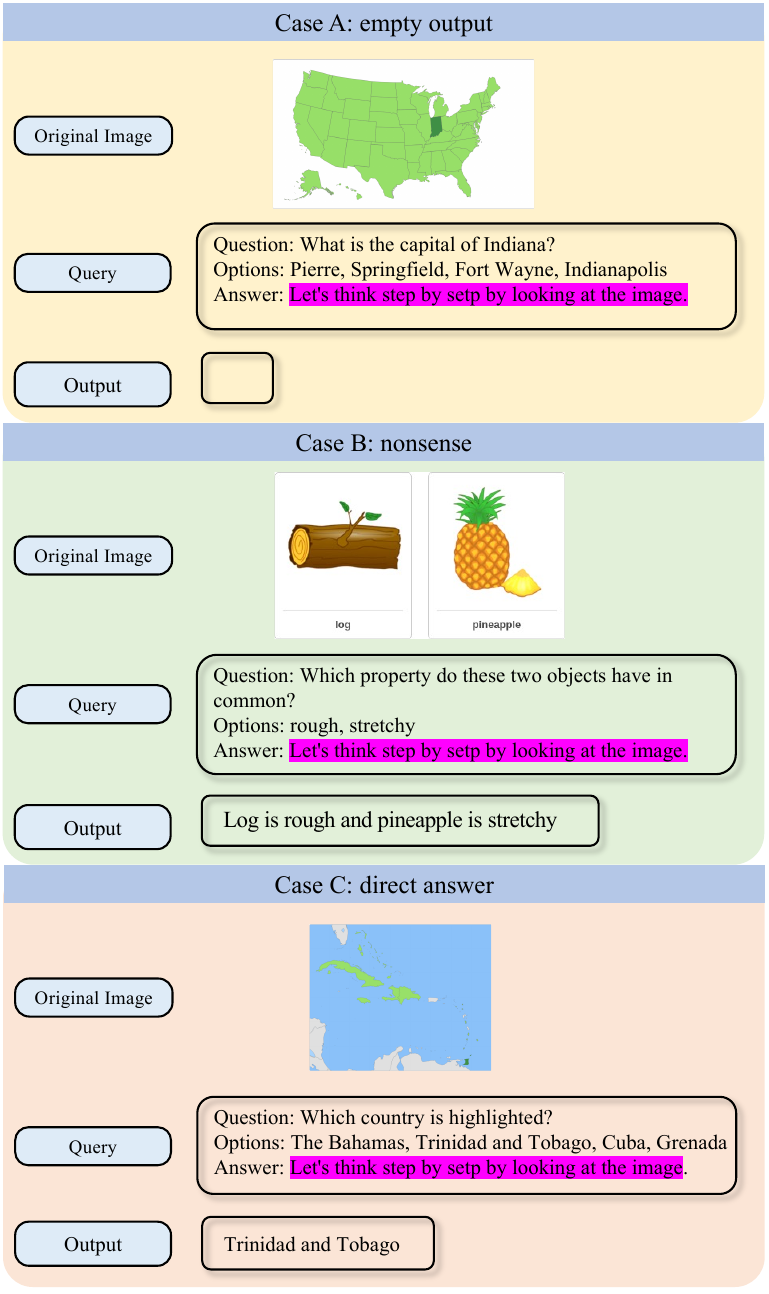}

   \caption{\footnotesize Zero-shot CoT generated by InstructBLIP.}
   \label{fig:cot}
   \vspace{-1em}
\end{figure*}

\section{Additional Analysis of QVix}
\label{sec:case study}

\subsection{Detailed Case Study of QVix}
QVix can utilize \textbf{detailed information} to better distinguish between options that are easily confused, and achieve a more comprehensive and systematic understanding of images through \textbf{contextual information}.

\paragraph{Detailed Information: Miniature Pinscher vs. Chihuahua}
The Miniature Pinscher and the Chihuahua are two breeds of dogs that look very similar. As shown in Fig.~\ref{fig:case detail}, the baseline InstructBLIP model often incorrectly identifies the Miniature Pinscher as a Chihuahua. The error arises because, merely through the short query, LVLM can not attend to the most crucial visual feature. On the contrary, by asking pre-question \textit{``Are there any specific facial features or characteristics that stand out on the pet in the image?''}, QVix attends to the eyes of the dog and thus correctly identifies the dog as Miniature Pinscher. This is because the eyes are a crucial detail in distinguishing between a Chihuahua and a Miniature Pinscher: the eyes of a Chihuahua are more protruding, while those of a Miniature Pinscher are flatter.

\paragraph{Contextual Information: Describe the system depicted in the image}
Contextual information can aid in a more comprehensive and systematic understanding of images. As illustrated in Fig.~\ref{fig:case context}, the query is: which sentence better describes the system depicted in the image. However, the baseline model's narrow focus on a restricted image region constrains its comprehension, resulting in an incomplete and partially correct response. On the contrary, by asking \textit{``What is the color of the water in the image?''} and \textit{``Is there any evidence of saltwater in the image, such as waves or sea spray?''} for sample A, and \textit{``What types of organisms can
be seen in the image?''} for sample B, QVix focuses on the contextual information within the image, thereby achieving a more comprehensive and holistic understanding of the image, which enables it to output a complete answer.

\subsection{QVix applied to GPT-4V}
\label{sec:gpt4v}

GPT-4V is one of the most powerful and recent LVLM. However, since it is close-sourced, we are unable to conduct quantitative experiments and are limited to performing qualitative experiments. The results are shown in Fig.~\ref{fig:gpt 1} and Fig.~\ref{fig:gpt 2}, where we highlight the wrong answer (red), correct answer (green), and rationale (blue) leading to the correct answer. 

In Fig.~\ref{fig:gpt 1}, QVix effectively resolves the hallucination issue of GPT-4V. Initially, GPT-4V incorrectly identifies the image as ``Air Jordan''. By asking GPT-4V the pre-questions \textit{``Is there a distinctive wing-like design or element in the image that could
be associated with the Air Jordan brand?''} and \textit{``Are there any numbers or letters in the image that correspond to a specific Air Jordan model or edition?''} generated by QVix, GPT-4V confirms that there are no definitive clues pointing to ``Air Jordan'' in the image, and thus provided the correct answer.

In Fig.~\ref{fig:gpt 2}, like InstructBLIP, GPT-4V fails to distinguish between Miniature Pinscher and Chihuahua. However, By asking GPT-4V the pre-questions \textit{``Are there any specific facial features or characteristics that stand out on the
pet in the image''}, GPT-4V generates the rationale \textit{``The muzzle is relatively short
with a well-defined stop, and the overall facial structure is characteristic of breeds like the
Miniature Pinscher''} which leads to the correct answer.

\subsection{Cost Analysis}
The additional cost of QVix comes from the generation of pre-questions.
For Fine-grained image classification and Visual Entailment, since we generate a set of pre-questions for each dataset and can reuse them across multiple instances, the cost is neglectable. For ScienceQA, we need to generate pre-questions for each instance first, and then append these pre-questions to the original query to perform the final reasoning. The cost in this case is close to the cost of zero-shot CoT, given that both approaches require on-the-fly generation of context-specific prompts or rationales for each individual instance.

\subsection{Randomness Analysis}
The only randomness in our method stems from invoking the GPT Api to generate pre-questions. We analyze the randomness of QVix in Table~\ref{tab:random}. We can see that the randomness in QVix is very low, with a small standard deviation, indicating that the improvements it brings are stable and robust.

\section{Analysis of Zero-shot CoT}
\label{sec:cot}
We analyze zero-shot CoT for LVLM. The prompt used to generate rationales is \textit{``Let's think step by step by looking at the image''}. In the experiments, LVLM fails to generate meaningful and informative rationales. As shown in Fig.~\ref{fig:cot}, the rationales generated by LVLM can be categorized into three types: the first yields empty responses; the second comprises incoherent rationales that contribute no valuable insight; and the third entirely bypasses the reasoning process, attempting to answer the original question directly.

\section{GPT-rewritten Pre-question Generation Prompt}
In section~\ref{sec:analysis}, we use GPT to rewrite our manully designed prompt in Fig.~\ref{fig:prompt}.
The different versions of GPT-rewritten pre-question generation prompt are shown in Fig.~\ref{fig:gpt re1} to Fig.~\ref{fig:gpt re3}. We can observe that the GPT-rewritten prompts do not alter the semantics of our manually designed prompts.

\begin{figure*}
\begin{center}
\fcolorbox{black}{gray!10}{\parbox{.9\linewidth}{I require assistance in formulating a response to a central inquiry regarding a specific image:  
\\ \hspace*{\fill} \\
\{Task Instruction\}
\\ \hspace*{\fill} \\
\{Query\}
\\ \hspace*{\fill} \\
The task is to create 4 preliminary questions. These questions should zero in on crucial contextual details within the image that are pertinent to addressing the main inquiry.
\\ \hspace*{\fill} \\
Guidelines for the preliminary questions:

Each question must be concise and easily comprehensible.
They should concentrate on contextual visual elements present in the image.
These questions ought to offer insights that aid in responding to the main question.
\\ \hspace*{\fill} \\
Proposed Format:

Preliminary Question 1: xxxx

Preliminary Question 2: xxxx

Preliminary Question 3: xxxx

Preliminary Question 4: xxxx
}}
\end{center}
   \vspace{-1.em}
 \caption{\footnotesize GPT-rewritten pre-question generation prompt version 1.}
 \label{fig:gpt re1}
    \vspace{-2.em}
\end{figure*}

\begin{figure*}
\begin{center}
\fcolorbox{black}{gray!10}{\parbox{.9\linewidth}{I am tasked with addressing a primary inquiry regarding a specific image:  
\\ \hspace*{\fill} \\
\{Task Instruction\}
\\ \hspace*{\fill} \\
\{Query\}
\\ \hspace*{\fill} \\
My objective is to formulate 4 preliminary questions. These questions are aimed at eliciting critical contextual details from the image, which are pivotal for comprehensively responding to the main inquiry.
\\ \hspace*{\fill} \\
Guidelines for crafting the preliminary questions:

Each question must be concise and easily comprehensible.
The focus should be on discerning visual cues within the image that offer context.
These questions are intended to unearth insights that facilitate answering the main question.
\\ \hspace*{\fill} \\
Formatted Example:

Preliminary Question 1: [Your Question Here]

Preliminary Question 2: [Your Question Here]

Preliminary Question 3: [Your Question Here]

Preliminary Question 4: [Your Question Here]

}}
\end{center}
   \vspace{-1.em}
 \caption{\footnotesize GPT-rewritten pre-question generation prompt version 2.}
 \label{fig:gpt re2}
    \vspace{-2.em}
\end{figure*}

\begin{figure*}
\begin{center}
\fcolorbox{black}{gray!10}{\parbox{.9\linewidth}{I am required to address a primary inquiry related to a specific image:  
\\ \hspace*{\fill} \\
\{Task Instruction\}
\\ \hspace*{\fill} \\
\{Query\}
\\ \hspace*{\fill} \\
The task is to formulate 4 preliminary questions. These questions are intended to extract key contextual details from the image that are crucial for responding accurately to the primary inquiry.
\\ \hspace*{\fill} \\
Guidelines for creating the preliminary questions:

Each question should be concise and straightforward for ease of understanding.
The focus should be on discernible contextual elements within the image.
These questions should aid in gathering insights necessary to address the primary inquiry.
\\ \hspace*{\fill} \\
Example of the format:

Preliminary Question 1: [Question text]

Preliminary Question 2: [Question text]

Preliminary Question 3: [Question text]

Preliminary Question 4: [Question text]
}}
\end{center}
   \vspace{-1.em}
 \caption{\footnotesize GPT-rewritten pre-question generation prompt version 3.}
 \label{fig:gpt re3}
    \vspace{-2.em}
\end{figure*}


\end{document}